\documentclass[]{spie}  

\usepackage{amsmath,amsfonts,amssymb}
\usepackage{graphicx}
\usepackage[colorlinks=true, allcolors=blue]{hyperref}
\usepackage{booktabs}
\usepackage{adjustbox}
\usepackage{pifont}
\newcommand{\cmark}{\ding{51}}%
\newcommand{\xmark}{\ding{55}}%

\newcommand{\etal}{\textit{et al.}}
\usepackage[table]{xcolor}

\title{Instance-Guided Radar Depth Estimation for 3D Object Detection}

\author{Chen-Chou Lo}
\author{Patrick Vandewalle}
\affil{EAVISE-PSI, Dept.\ of Electrical Engineering (ESAT), KU Leuven \\
Jan de Nayerlaan 5, 2860 Sint-Katelijne-Waver, Belgium}

\authorinfo{Further author information: chenchou.lo@kuleuven.be; patrick.vandewalle@kuleuven.be}

\begin{document} 
\maketitle

\begin{abstract}

Accurate depth estimation is fundamental to 3D perception in autonomous driving, supporting tasks such as detection, tracking, and motion planning. However, monocular camera-based 3D detection suffers from depth ambiguity and reduced robustness under challenging conditions. Radar provides complementary advantages such as resilience to poor lighting and adverse weather, but its sparsity and low resolution limit its direct use in detection frameworks. This motivates the need for effective Radar–camera fusion with improved preprocessing and depth estimation strategies.

We propose an end-to-end framework that enhances monocular 3D object detection through two key components.
First, we introduce InstaRadar, an instance segmentation–guided expansion method that leverages pre-trained segmentation masks to enhance Radar density and semantic alignment, producing a more structured representation. InstaRadar achieves state-of-the-art results in Radar-guided depth estimation, showing its effectiveness in generating high-quality depth features.
Second, we integrate the pre-trained RCDPT into the BEVDepth framework as a replacement for its depth module. With InstaRadar-enhanced inputs, the RCDPT integration consistently improves 3D detection performance.
Overall, these components yield steady gains over the baseline BEVDepth model, demonstrating the effectiveness of InstaRadar and the advantage of explicit depth supervision in 3D object detection.
Although the framework lags behind Radar–camera fusion models that directly extract BEV features, since Radar serves only as guidance rather than an independent feature stream, this limitation highlights potential for improvement. Future work will extend InstaRadar to point cloud–like representations and integrate a dedicated Radar branch with temporal cues for enhanced BEV fusion.

\end{abstract}


\section{Introduction}
\label{sec:3dod_intro}

To enable autonomous driving, a comprehensive 3D understanding of the environment is essential. Depth estimation provides the initial perception, while 3D object detection underpins many downstream tasks. Among available sensors, RGB cameras are widely adopted due to their low cost, easy integration, and high-resolution visual information, making multi-view camera-based 3D perception increasingly popular. However, monocular depth estimation is inherently ill-posed, limiting accurate 3D reasoning. Early approaches extended 2D detectors with depth or orientation predictions, but lacked depth supervision. Pseudo-LiDAR methods~\cite{PseudoLiDAR} addressed this by projecting estimated depth maps into point cloud-like representations for LiDAR-based detectors, yet their performance remains sensitive to depth accuracy, with noisy estimates leading to degraded detection.

To address these limitations, two main directions have been explored. One is to explicitly supervise depth prediction during training, reducing noise in pseudo-LiDAR representations. The other is to directly learn the 2D-to-3D projection in an end-to-end manner. A representative example is Lift-Splat-Shoot (LSS) \cite{LSS}, which lifts image features into 3D frustum volumes using estimated depth and projects them onto a bird’s-eye-view (BEV) plane. Unlike pseudo-LiDAR, BEV-based methods alleviate projection errors and perspective distortion, yielding more consistent spatial representations. Since LSS, many works\cite{BEVDet4D, BEVDet, BEVDepth, BEVFormer,RCBEVDet,SparseBEV0} have adopted BEV frameworks for 3D object detection, achieving strong performance and becoming widely used.

Despite the progress of BEV-based methods, the lack of inherent depth in camera images remains a challenge. Integrating LiDAR can alleviate this issue, but it is expensive and weather-sensitive, limiting large-scale deployment. In contrast, Radar and camera sensors are cost-effective, robust under various conditions, and already widely used in production vehicles for safety applications. Cameras provide rich semantic information, while Radar offers long-range and weather-robust characteristics. These complementary properties have motivated research into combining Radar with camera-based BEV methods to enhance depth perception and 3D detection.
Rather than directly feeding raw Radar data into 3D models, Radar-guided depth estimation~\cite{MDE_radar, DORN_radar, RCDPT, JBF_radar} has demonstrated that preprocessed Radar depth can significantly improve depth precision. However, existing Radar preprocessing approaches often rely on simple expansion techniques~\cite{DORN_radar} or on training dedicated neural networks~\cite{RC_PDA}, which may not generalize well to unseen environments.

In this work, we introduce InstaRadar, an instance-guided Radar expansion method, and propose a 3D object detection framework that leverages this enhanced Radar representation together with a Radar-guided depth estimation model to improve detection performance.
Our contributions are twofold.
First, we introduce InstaRadar, an instance segmentation-guided expansion method that leverages OneFormer~\cite{OneFormer} to densify raw Radar points and align them with object-level semantics, increasing Radar coverage while preserving object-awareness.
Second, we integrate the pre-trained RCDPT\cite{RCDPT} into the BEVDepth\cite{BEVDepth} pipeline, replacing its default depth module. 
By fusing Radar and image features through a transformer-based cross-modal reassembly mechanism~\cite{DPT}, RCDPT explicitly supervises depth prediction and improves the quality of intermediate depth features.
The resulting depth maps are projected into BEV space via voxel pooling, producing more accurate spatial features for 3D detection. 
Overall, the framework demonstrates that combining object-aware Radar expansion with explicit Radar-guided depth supervision not only improves the baseline 3D object detection model but also achieves SOTA performance in depth estimation.


\section{Related Work}
\label{sec:3dod_related_work}

3D object detection aims to identify and localize objects in 3D space, which is essential for autonomous driving. Recent methods extend 2D detectors to predict 3D information from monocular or multi-view images~\cite{DETR3D}, but their accuracy is limited by the lack of depth cues. Pseudo-LiDAR approaches~\cite{PseudoLiDAR} convert estimated depth maps into point clouds for LiDAR-based pipelines, yet remain sensitive to depth quality. Alternatively, BEV-based methods~\cite{LSS} provide a unified and spatially consistent representation, making them well-suited for 3D perception.

\subsection{Camera-based 3D Object Detection in BEV}

A key development in 3D detection is the transformation of image features into the BEV format, where detection is performed from a top-down perspective. BEV representations offer a unified spatial coordinate system aligned with the ego-vehicle, reduce perspective distortion, and facilitate downstream tasks such as map segmentation and trajectory planning. BEV-based methods also enable flexible sensor fusion and allow efficient reasoning over spatial layouts. This direction has led to a series of advances in camera-based detection frameworks that leverage BEV representations for improved robustness and accuracy~\cite{BEVDet4D, BEVDet, BEVDepth, BEVFormer, RCBEVDet, SparseBEV0, LSS}.

Philion and Fidler~\cite{LSS} introduced LSS, which lifts image features into 3D frustums using predicted depth and projects them onto a BEV grid. This differentiable pipeline enables effective multi-camera BEV perception. Building on LSS, Huang \etal~\cite{BEVDet} proposed BEVDet, which enhances 3D detection with a modular architecture including a depth-aware view transformer and BEV encoder, and introduces Scale-NMS to better handle objects of varying sizes.
To improve temporal reasoning, BEVDet4D~\cite{BEVDet4D} extends BEVDet with motion-aware modules and temporal feature fusion based on sequences of camera images. BEVFormer~\cite{BEVFormer} replaces explicit depth estimation with a spatial and temporal transformer, using deformable attention to directly aggregate multi-view features in BEV space. SparseBEV~\cite{SparseBEV0} introduces a query-based sparse attention framework, avoiding dense BEV construction and improving both accuracy and efficiency.

Despite these developments, the accuracy of BEV-based methods is still limited by the quality of the estimated depth, which is often weakly supervised. To address this, BEVDepth~\cite{BEVDepth} introduces explicit depth supervision using sparse LiDAR, along with a camera-aware depth module and a refinement mechanism to improve feature alignment. These enhancements significantly improve the reliability of BEV features and overall detection performance.

\subsection{Camera-based 3D Object Detection with Radar}

While monocular BEV detectors show strong potential, they remain constrained by unreliable depth estimation and poor performance under adverse lighting or weather conditions. These limitations become more pronounced at long ranges, where monocular cues are insufficient. To address this, Radar sensors have been introduced as a complementary modality, offering robust long-range measurements and direct velocity estimation. By fusing Radar with camera images, models can combine the semantic richness of visual features with the spatial accuracy and robustness of Radar, improving overall detection reliability.

CRFNet~\cite{CRFNet} is one of the earliest Radar-camera fusion models on the nuScenes dataset, projecting Radar points onto the image plane and applying a dropout-like strategy to encourage meaningful Radar feature learning. CenterFusion~\cite{CenterFusion} uses a frustum-based association method to fuse Radar detections with image-based object centers, enhancing depth and velocity estimation. HVDetFusion~\cite{HVDetFusion} extends BEVDet4D with a separate Radar branch that filters noisy points using BEV priors and fuses the Radar features with image features for refined predictions. RCBEVDet~\cite{RCBEVDet} improves BEV-based Radar fusion through a Radar feature extraction module and deformable cross-attention, enabling better alignment between Radar and image features in BEV space.
CRN~\cite{CRN} introduces a two-stage Radar-camera fusion framework that generates accurate and semantically rich BEV features. It leverages Radar-guided view transformation using Radar occupancy as spatial priors to enhance the projection of image features, and then applies cross-attention for multi-modal feature fusion while addressing misalignment and Radar sparsity. CRN demonstrates SOTA performance across detection, tracking, and segmentation tasks on nuScenes, showing strong robustness and real-time efficiency.

The reviewed works highlight that the quality of estimated depth is critical for BEV-based 3D object detection. Inaccurate intermediate depth leads to erroneous BEV feature transformations and degraded predictions. Prior studies~\cite{MDE_radar, LO_EI2023, RC_PDA, RCDPT, JBF_radar} have shown that incorporating Radar into depth estimation can improve accuracy, and that appropriate Radar preprocessing enhances resolution and feature extraction. Motivated by these findings, we integrate a Radar-guided depth estimation module into a 3D detection framework to refine intermediate depth and thereby improve final predictions. In addition, we propose an instance-aware Radar expansion method to increase Radar resolution and strengthen feature extraction.

\section{Method}

In this section, we introduce two key components of our methodology. First, we present InstaRadar, an instance segmentation-guided expansion that enhances raw Radar resolution using instance masks. Second, we describe the integration of a Radar-guided depth estimation module into a 3D detection framework to improve intermediate depth features and overall detection performance.

\subsection{Instance Segmentation-guided Radar Expansion}


\begin{figure}
\centering
\includegraphics[width=\textwidth,height=\textheight,keepaspectratio]{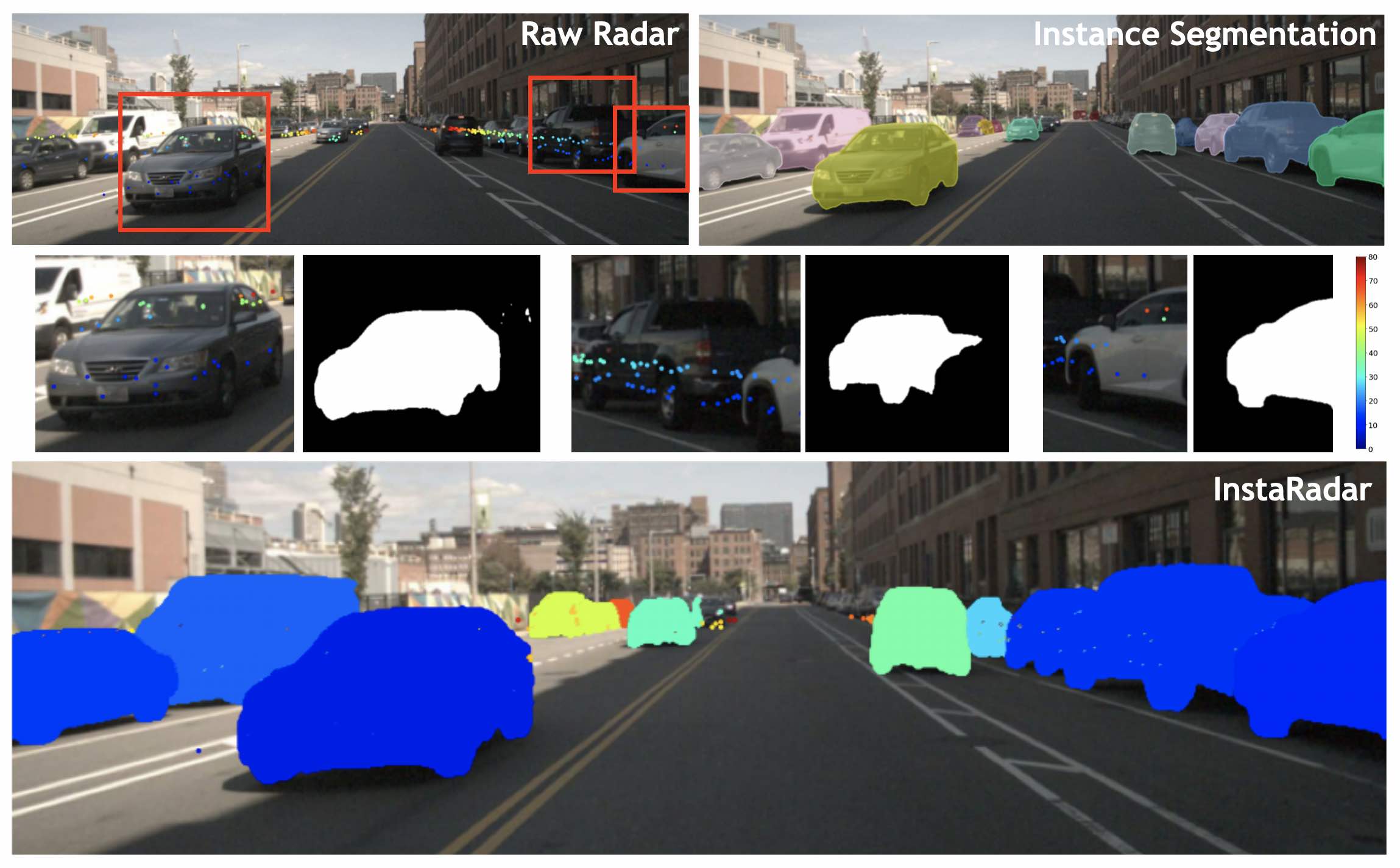}
\caption[Visualization of raw Radar, instance segmentation, instance masks, and the proposed Instance Segmentation-guided Radar Expansion (InstaRadar).]
{
Visualization of raw Radar, instance segmentation, instance masks, and the proposed InstaRadar on the nuScenes dataset~\cite{nuScenes}.
The top-left shows raw Radar points projected from 5 frames onto the reference image, while the top-right displays instance segmentation results from OneFormer~\cite{OneFormer}. The middle row highlights close-up views of the Radar depth points and corresponding instance masks from the boxes in the top-left image. The bottom image presents the final InstaRadar result, where Radar points are expanded within instance masks to improve resolution and spatial coverage. Radar points are enlarged for better visibility, and the color bar indicates distances ranging from 0 to 80 meters.
} 
  \label{fig:3dod_insta_radar_method}
\end{figure}


Previous studies have shown that properly preprocessed Radar data can significantly enhance depth estimation in Radar-guided models, leading to better performance in 3D object detection. Existing Radar expansion methods include relatively simple techniques such as two-stage filtering~\cite{MDE_radar}, height-based extension~\cite{DORN_radar}, or using a pre-trained network~\cite{Singh} or even under LiDAR supervision ~\cite{RC_PDA}. However, these approaches often fail to leverage rich visual information from images, and pre-trained models typically require fine-tuning with LiDAR data, limiting their generalization to unseen environments. 
In contrast, the JBF-based Radar expansion method~\cite{JBF_radar} employs image information as a reference through a joint bilateral filter to guide Radar point expansion. While it improves spatial resolution, this design suffers from fixed kernel sizes that limit coverage, requires manual tuning of kernel parameters, and represents objects only at the pixel level, leading to poor performance when object variations are large.

To address these limitations, we propose instance segmentation-guided Radar expansion, InstaRadar, which leverages instance masks to guide the expansion process. A SOTA instance segmentation model, OneFormer~\cite{OneFormer}, is adopted to obtain object-level masks. This enables the preservation of object structures during expansion and generates more localized and meaningful Radar features for downstream tasks.
Examples of raw Radar, instance segmentation, and the proposed InstaRadar on the nuScenes dataset~\cite{nuScenes} are shown in \autoref{fig:3dod_insta_radar_method}. The top-left image illustrates raw Radar points projected from 5 consecutive frames onto the current image, while the top-right shows instance segmentation results obtained using OneFormer~\cite{OneFormer}. Leveraging individual instance masks, the Radar depth values within each object region are examined, as shown in the middle row. 
Within a single object mask, depth variations may arise from sensor noise or object size. Significant depth differences are often due to occlusion.
Since Radar points are aggregated from previous frames, distant objects may be partially occluded by nearer ones. In such cases, the closest depth value is assigned as dominant, assuming that the nearer object is in front. After expanding Radar points within all detected instance masks, the expanded Radar depth is overlaid with the original ground truth Radar depth to produce the final result. Following this procedure, the proposed InstaRadar is generated, as shown in the bottom image.

\begin{figure}
\centering
\includegraphics[width=\textwidth,height=\textheight,keepaspectratio]{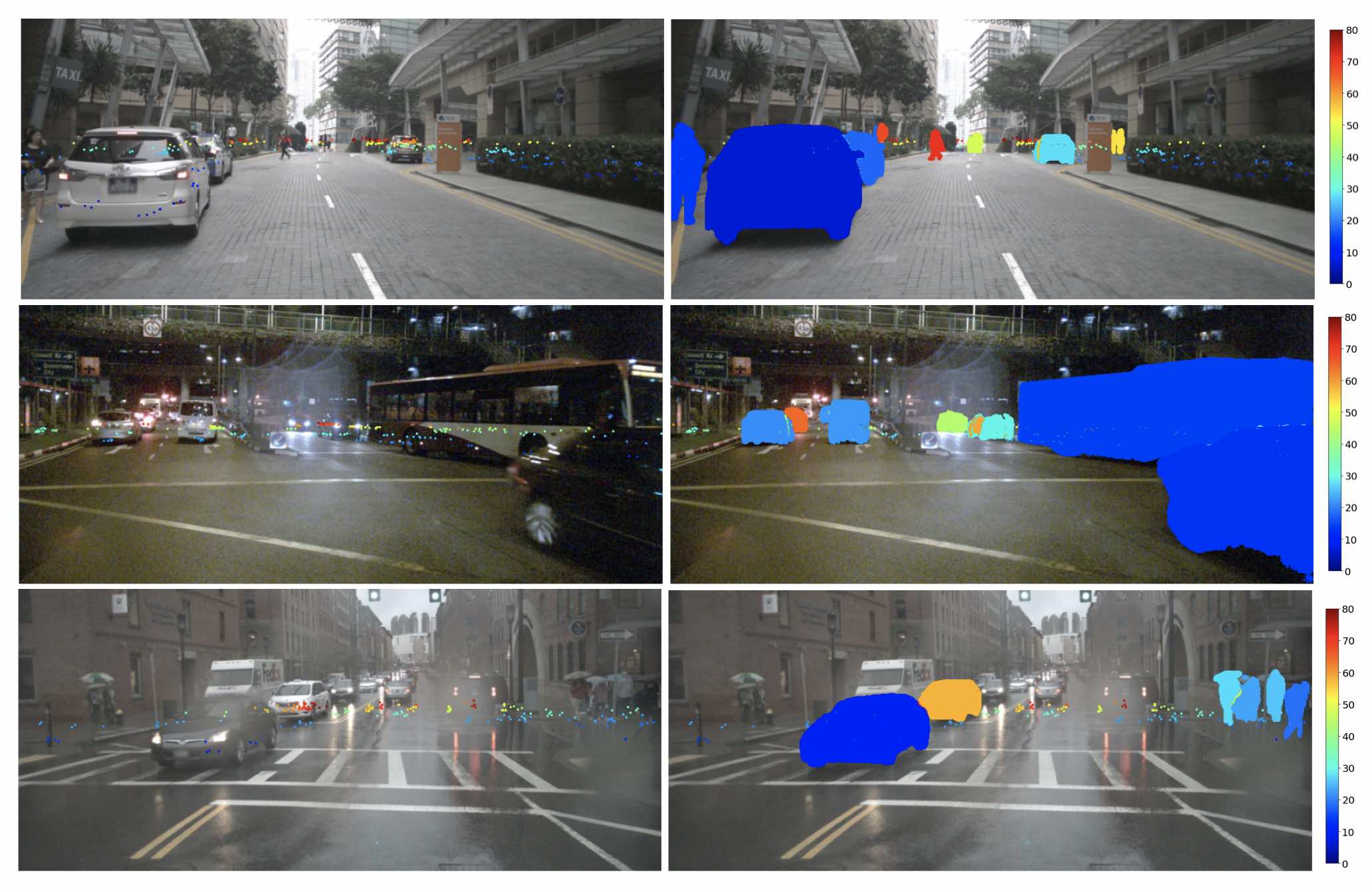}

  \caption[Examples of the proposed Instance Segmentation-guided Radar Expansion]
  {
  Examples of the proposed Instance Segmentation-guided Radar expansion on the nuScenes dataset~\cite{nuScenes}.
  The left column shows raw Radar points from 5 frames, and the right column shows InstaRadar expanded within instance masks.
  Rows from top to bottom correspond to day, night, and rain scenes. Radar points are enlarged for clarity, with distances indicated by the 0–80 m color bar.
  } 
  \label{fig:3dod_insta_radar_samples}
\end{figure}

Additional examples of the proposed InstaRadar are shown in \autoref{fig:3dod_insta_radar_samples}, illustrating qualitative results under different lighting and weather conditions, including day, night, and rain scenes. In each pair, the left column shows the camera image with raw Radar depth overlaid, while the right column illustrates the proposed InstaRadar. It can be observed that the instance masks help Radar depth align well with object regions, allowing for more accurate depth coverage. 
Across various environmental conditions, InstaRadar consistently captures objects such as vehicles and pedestrians, demonstrating its robustness and ability to generate semantically meaningful Radar representations for downstream perception tasks.


\subsection{Radar-guided Depth Estimation Integrated into 3D Object Detection}


\begin{figure}
\centering
\includegraphics[width=\textwidth,height=\textheight,keepaspectratio]{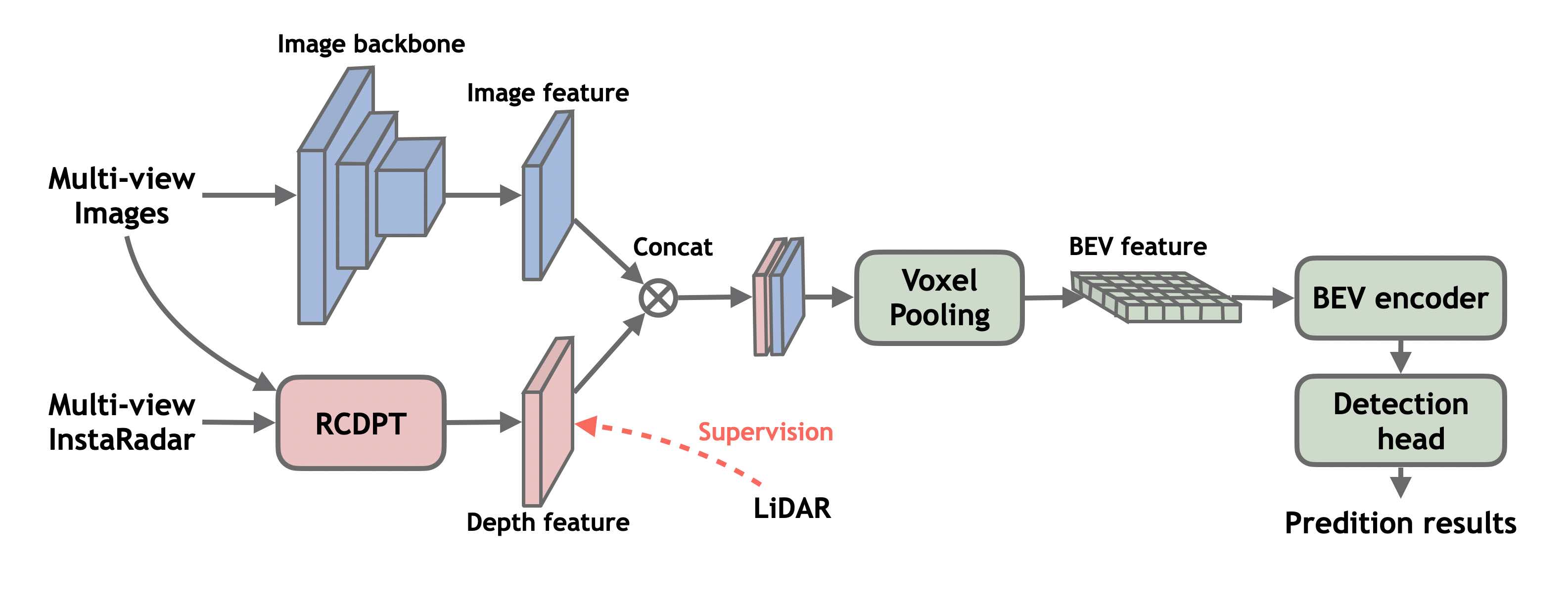}

  \caption[Framework of the proposed method]
  {
  Our proposed framework takes multi-view images and InstaRadar as input. The image backbone extracts image features, while a pre-trained RCDPT takes both image and InstaRadar as input to predict depth features under LiDAR supervision. These image and depth features are concatenated and passed through voxel pooling to generate a BEV feature map. Finally, the BEV feature map is passed through a BEV encoder and a detection head to output the final 3D detection results.
  } 
  \label{fig:3dod_proposed_3dod}
\end{figure}


In addition to the proposed Radar preprocessing method, we further improve the depth feature map in the intermediate layers by integrating a Radar-guided depth estimation module into an existing BEV-based 3D object detection model. As shown in BEVDepth \cite{BEVDepth}, although depth can be implicitly learned during end-to-end training in a BEV-based model, the quality of the estimated depth is often vague and lacks spatial precision. To tackle this, BEVDepth introduces explicit depth supervision using ground truth LiDAR depth along with an additional depth loss, which significantly improves the performance of detection.
Unlike the default depth estimation module in BEVDepth, RCDPT is specifically designed to leverage Radar data for more accurate depth prediction based on a vision transformer structure. 
In our method, the default depth module is replaced with a pre-trained RCDPT model, enabling direct integration of Radar depth into the 3D detection pipeline.

\autoref{fig:3dod_proposed_3dod} illustrates the overall architecture of the proposed system. 
The framework takes multi-view images and Radar as input to predict accurate 3D object detections in the BEV space. 
First, the image backbone (shown in blue) processes the multi-view images to extract image features.
In parallel, the pre-trained RCDPT (shown in red) takes both the multi-view images and the corresponding InstaRadar as input to generate intermediate depth features. This depth estimation process is fine-tuned with LiDAR supervision during training.
The extracted image features and depth features are then concatenated to form a unified representation, which is passed through a voxel pooling layer to project the features into the BEV space.
Finally, the BEV feature map (shown in green) is processed by a BEV encoder, followed by a detection head, to produce the final 3D detection results.

One of the key novelties of the proposed framework is that the integrated RCDPT is fed with enhanced Radar depth generated by the proposed InstaRadar method.
By leveraging instance-guided Radar expansion, InstaRadar provides a denser and more structured perception, enabling RCDPT to generate more reliable depth features across objects in the scene.
This integration improves the quality of the estimated depth in the intermediate layers and BEV space, leading to better overall 3D object detection performance.
The results demonstrate that combining a strong Radar-guided depth estimation model with an effective Radar preprocessing method, such as InstaRadar, can significantly enhance BEV-based detection frameworks.


\section{Experiments}
\label{sec:3dod_experiments}

\subsection{Experimental Setup}

{\bf Dataset and metrics.}
We conduct our experiments on the nuScenes dataset~\cite{nuScenes}, a large-scale multimodal benchmark for autonomous driving that provides synchronized data from 6 cameras, 5 Radars, and a 32-beam Velodyne LiDAR. It contains 1,000 driving scenes, divided into 700 for training, 150 for validation, and 150 for testing, with each scene comprising 40 key-frame samples recorded over 20 seconds. 
For depth estimation, we report RMSE and AbsRel metrics following commonly used measures in prior works~\cite{MDE_radar, DORN_radar, JBF_radar}. 
For 3D object detection, we report mAP and NDS, consistent with previous methods~\cite{BEVDepth, BEVDet, CRN, CenterFusion} on the nuScenes dataset.

\noindent
{\bf Implementation details.}
To generate InstaRadar, we first apply a pre-trained SOTA instance segmentation model, OneFormer~\cite{OneFormer}, to all camera views in the nuScenes dataset to obtain instance masks. Using these masks, our Radar expansion method assigns a dominant depth to each segmented region by selecting the nearest Radar point, yielding a denser Radar representation termed InstaRadar. We then pre-train RCDPT~\cite{RCDPT} with InstaRadar as an additional Radar input across all camera views, following the training procedure and settings in the RCDPT paper~\cite{RCDPT}.
For 3D object detection, we adopt BEVDepth \cite{BEVDepth} as the baseline and replace its default depth module with the pre-trained RCDPT to enhance intermediate depth quality. 
All models are trained for 20 epochs using the AdamW~\cite{AdamW} optimizer on a single NVIDIA V100 GPU, with a learning rate of 2e-4 and a batch size of 16. Image augmentation includes random horizontal flipping and cropping 900$\times$1600 images to 256$\times$704. In BEV space, features and detection targets are further augmented with random flipping, rotation within ±22.5°, and scaling from 0.95 to 1.05.

\subsection{Radar-guided Depth Estimation with InstaRadar}


\begin{table}[ht]
\centering
\caption[Radar-guided depth estimation results.]
{
Radar-guided depth estimation results.
}
\label{tab:3dod_depth_result}
\begin{tabular}{lcccc}
\toprule
\multicolumn{1}{c}{\textbf{Method}} & \textbf{Supervision} & \textbf{Radar} & \textbf{AbsRel $\boldsymbol{\downarrow}$} & \textbf{RMSE $\boldsymbol{\downarrow}$} \\ 
\midrule
BEVDepth (default) & \xmark & Raw & 3.03 & 19.45 \\
BEVDepth (default)  & \cmark & Raw & 0.23 & 5.78 \\
\midrule
S2D~\cite{s2d} & \cmark & Raw & 0.12 & 5.63 \\
$DORN_{radar}$~\cite{DORN_radar} & \cmark & Height-extended & 0.11 & 5.26 \\
RCDPT& \cmark & MER & 0.10 & 5.16 \\
RCDPT & \cmark & JBF & 0.09 & 4.88 \\
\midrule
\rowcolor{gray!20} RCDPT & \cmark & InstaRadar & 0.09 & 5.09 \\
\bottomrule
\end{tabular}
\end{table}



We first evaluate the proposed InstaRadar for depth estimation, comparing it with the default depth module in BEVDepth. Results are shown in \autoref{tab:3dod_depth_result}. The top two rows report the original BEVDepth results~\cite{BEVDepth}, while the remaining rows include prior methods and our experiments. The supervision column indicates whether LiDAR supervision is used during training.

The first row in the table shows the default depth module of BEVDepth trained without LiDAR supervision, similar to the implicit depth estimation in LSS or BEVDet. As expected, LiDAR supervision greatly improves depth quality, but a gap remains compared to models specifically designed for depth estimation.
Our proposed InstaRadar, trained with RCDPT, outperforms the default module in both AbsRel and RMSE and achieves competitive performance compared to SOTA JBF Radar. While InstaRadar improves over the baseline, it falls short of JBF Radar in RMSE, possibly because JBF expansion exploits pixel-wise similarity across the image to produce denser depth in background and far-distance regions. In contrast, InstaRadar is object-aware, guided by instance segmentation, and concentrates Radar expansion on foreground objects. This yields stronger semantic alignment but less coverage at long range, where larger errors have a greater effect on RMSE. 
Nevertheless, InstaRadar provides clear advantages in preserving object-level consistency and generating instance-aware depth, which are more valuable for downstream 3D object detection. 
These results highlight the benefit of combining high-quality Radar input from InstaRadar with a Radar-guided model like RCDPT.

\subsection{Radar-guided 3D Object Detection}

\subsubsection{Baseline reproduction and Radar-guided depth}

To establish a reliable baseline, we examine the reproducibility of BEVDepth, since our method builds on its architecture by replacing the depth module with a pre-trained Radar-guided network.

\autoref{tab:3dod_reproduce_bevdetph} summarizes the comparison of different BEVDepth versions. The first row shows the original results reported in the BEVDepth paper~\cite{BEVDepth}, which achieved the highest mAP (0.351) and NDS (0.475). Subsequent rows present reproductions from other works~\cite{BEVStereo, bevdepth_r2, bevdepth_r1}, with mAP values around 0.33 and NDS between 0.40 and 0.43, highlighting the difficulty of matching the original performance. Our reproduced version, despite closely following the official training settings, achieves 0.336 mAP and 0.416 NDS, consistent with these reports. We therefore adopt it as the baseline for subsequent experiments.

The last three rows of the table show our ablation studies on the BEVDepth architecture with JBF Radar and InstaRadar inputs. 
Replacing raw Radar with either format yields small but consistent gains in mAP and NDS, indicating that both provide richer depth information. 
Interestingly, although InstaRadar performed slightly worse than JBF Radar in depth estimation, it achieves better results in 3D detection, suggesting that InstaRadar contributes more meaningful structural cues for object understanding. 
Finally, replacing the default depth module with our pre-trained RCDPT using InstaRadar input leads to a significant boost, reaching 0.355 mAP (+5.65\%) and 0.457 NDS (+9.86\%) over the baseline. These results confirm the effectiveness of our approach in enhancing intermediate depth estimation and overall 3D object detection.


\begin{table}
\centering
\caption[Comparison of BEVDepth reproduced variants and proposed Radar-guided extensions on the nuScenes val set.]
{
Comparison of BEVDepth reproduced variants and proposed Radar-guided extensions on the nuScenes \textbf{val} set.
}
\label{tab:3dod_reproduce_bevdetph}
\begin{tabular}{lcccc}
\toprule
\multicolumn{1}{c}{\textbf{Method}} & \textbf{Depth Backbone} & \textbf{Radar} & \textbf{mAP $\boldsymbol{\uparrow}$} & \textbf{NDS $\boldsymbol{\uparrow}$} \\ 
\midrule
BEVDepth-R50~\cite{BEVDepth} & default depth module & Raw & 0.351 & 0.475 \\
\midrule
BEVDepth-R50~\cite{BEVStereo} & default depth module & Raw & 0.327 & 0.433 \\
BEVDepth-R50~\cite{bevdepth_r2} & default depth module & Raw & 0.333 & 0.406 \\
BEVDepth-R50~\cite{bevdepth_r1} & default depth module & Raw & 0.332 & 0.404 \\
\midrule
BEVDepth-R50 (Ours) & default depth module & Raw & 0.336 & 0.416 \\
BEVDepth-R50 (Ours) & default depth module & JBF & 0.338 & 0.420 \\
BEVDepth-R50 (Ours) & default depth module & InstaRadar & 0.339 & 0.422\\
\rowcolor{gray!20} Proposed & RCDPT & InstaRadar & 0.355 & 0.457 \\

\bottomrule
\end{tabular}
\end{table}


\subsubsection{Comparison with monocular and Radar–camera fusion 3D detection methods}

\autoref{tab:3dod_3dod_result} compares our method with representative 3D object detection models on the nuScenes validation set, including LiDAR, monocular, and Radar–camera-based approaches. The top rows list classical LiDAR models as references, followed by monocular image-based methods in the middle. The bottom rows show our proposed method alongside other Radar–camera fusion models. For fairness, we report versions with comparable image resolution (256$\times$704) and ResNet-50 backbones where applicable.

For monocular camera-based methods, our approach achieves the highest mAP (0.355) and NDS (0.457). Compared to BEVDepth~\cite{BEVDepth}, BEVDet4D~\cite{BEVDet4D}, PETR~\cite{PETR}, and BEVDet~\cite{BEVDet}, it benefits from Radar-guided depth estimation and InstaRadar, which enhance depth quality and semantic consistency, leading to improved 3D localization and more robust detection. Although LiDAR-based models such as VoxelNet~\cite{VoxelNet} and PointPillars~\cite{PointPillars} still achieve the highest overall performance, our method provides a competitive and cost-effective alternative without relying on expensive LiDAR. These results show that combining camera and Radar with improved depth features can substantially enhance detection accuracy and represent promising directions for monocular 3D object detection.


\begin{table}
\centering
\caption[3D object detection result on the nuScenes val set]
{
3D object detection result on the nuScenes \textbf{val} set. In the Modality column, L denotes LiDAR, C denotes Camera, and C+R denotes Camera-Radar fusion methods.
}
\label{tab:3dod_3dod_result}
\begin{tabular}{lcccc}
\toprule
\multicolumn{1}{c}{Method} & Image Size & Modality & \textbf{mAP $\boldsymbol{\uparrow}$} & \textbf{NDS $\boldsymbol{\uparrow}$} \\ 
\midrule 
VoxelNet~\cite{VoxelNet} & - & L & 0.563 & 0.648 \\
PointPillar~\cite{PointPillars} & - & L & 0.487 & 0.597 \\
\midrule 
CenterNet~\cite{CenterNet} & 900$\times$1600 & C & 0.328 & 0.306 \\
DETR3D~\cite{DETR3D} & 900$\times$1600 & C & 0.303 & 0.374 \\
BEVDet-R50~\cite{BEVDet} & 256$\times$704 & C & 0.298 & 0.379 \\
PETR-R50~\cite{PETR} & 384$\times$1056 & C & 0.313 & 0.381 \\
BEVDet4D-Tiny~\cite{BEVDet4D} & 256$\times$704 & C & 0.323 & 0.453 \\
BEVDepth-R50~\cite{BEVStereo} & 256$\times$704 & C & 0.327 & 0.433 \\
\midrule
CenterFusion~\cite{CenterFusion} & 450$\times$800 & C+R & 0.332 & 0.453 \\

CRAFT~\cite{CRAFT} & 448$\times$800 & C+R & 0.411 & 0.517 \\

CRN-R50~\cite{CRN} & 448$\times$800 & C+R & 0.490 & 0.560 \\

\rowcolor{gray!20} Proposed & 256$\times$704 & C+R & 0.355 & 0.457 \\

\bottomrule
\end{tabular}
\end{table}


However, our method outperforms all monocular models but still lags behind dedicated fusion architectures when compared with Radar–camera fusion approaches. This is mainly because, although our framework integrates Radar as an additional input, it primarily enhances intermediate depth prediction within a monocular pipeline. In contrast, other Radar–camera fusion models~\cite{CRAFT, CRN} process Radar in point cloud format to generate BEV features, which are then fused with image BEV features through specialized fusion structures. By exploiting Radar’s spatial precision and motion cues, these models achieve stronger performance.

Although our model consistently improves over its baseline, it has yet to surpass these advanced Radar–camera fusion approaches. Nonetheless, their architectures suggest promising directions for future work. In particular, extending InstaRadar into a point cloud representation and incorporating a dedicated Radar branch for BEV fusion could enable richer multimodal representations and further boost 3D detection performance.

\section{Conclusion}
\label{sec:3dod_conclusion}

In this work, we proposed a novel 3D object detection framework that enhances monocular depth estimation through two key contributions. First, we introduced InstaRadar, an instance segmentation-guided Radar expansion method that improves the resolution and object-level alignment of Radar data. InstaRadar achieves competitive performance on the nuScenes validation set for Radar-guided depth estimation, offering stronger object-aware localization and semantic consistency that are particularly beneficial for detection tasks. Second, we replaced the default depth module of BEVDepth with the RCDPT model, which explicitly supervises depth estimation and improves the quality of intermediate depth features. Together, these components yield consistent improvements over the baseline BEVDepth model, highlighting both the effectiveness of InstaRadar and the benefit of explicit depth supervision with a dedicated depth module in 3D object detection frameworks.

Despite these improvements, the framework still lags behind stronger Radar-camera fusion methods, as Radar is used mainly as depth guidance rather than as an independent feature stream. Future work includes incorporating temporal information and extending InstaRadar into a BEV representation with a dedicated Radar branch, enabling richer multimodal fusion and narrowing the gap with advanced fusion models.

 

\bibliography{references} 
\bibliographystyle{spiebib} 

\end{document}